\documentclass[runningheads,a4paper]{llncs}
\usepackage[T1]{fontenc}
\usepackage{amsmath}
\usepackage{graphicx}
\usepackage[colorlinks=true, allcolors=blue]{hyperref}
\usepackage{amssymb} % or amsfonts
\usepackage{url}
\usepackage{subfig}
\newcommand{\reels}{\mathbb{R}}
\newcommand{\tT}{\widetilde{T}}
\newcommand{\tN}{\widetilde{N}}
\newcommand{\tF}{\widetilde{F}}
\newcommand{\tX}{\widetilde{X}}
\newcommand{\tY}{\widetilde{Y}}
\newcommand{\tpsi}{\widetilde{\psi}}

\newcommand{\calL}{{\cal L}}

\begin{document}
\title{An evidential time-to-event prediction model based on Gaussian random fuzzy numbers}
\titlerunning{Time-to-event prediction model based on Gaussian random fuzzy numbers}
\author{Ling Huang\inst{1} \orcidID{0000-0003-1562-1371}\and
Yucheng Xing\inst{1} \and
Thierry Den{\oe}ux\inst{3,4} \orcidID{0000-0002-0660-5436} \and
Mengling Feng\inst{1,2} \orcidID{0000-0002-5338-6248}}
\authorrunning{L. Huang et al.}
\institute{Saw Swee Hock School of Public Health, \\
National University of Singapore, Singapore \\
\email{huang.l@nus.edu.sg}
\and
Institute of Data Science, National University of Singapore, Singapore
\and
Université de technologie de Compiègne, CNRS, Heudiasyc, France \and
Institut Universitaire de France, France}
\maketitle              % typeset the header of the contribution
\begin{abstract}
We introduce an evidential model for time-to-event prediction with censored data. In this model, uncertainty on event time is quantified by Gaussian random fuzzy numbers, a newly introduced family of random fuzzy subsets of the real line with associated belief functions, generalizing both Gaussian random variables and Gaussian possibility distributions. Our approach makes minimal assumptions about the underlying time-to-event distribution. The model is fit by minimizing a generalized negative log-likelihood function that accounts for both normal and censored data. Comparative experiments on two real-world datasets demonstrate the very good performance of our model as compared to the state-of-the-art.

\end{abstract}
\section{Introduction}

Time-to-event analysis, also known as survival analysis, focuses on analyzing the time it takes for an event of interest to occur, such as time to death or machine failure. 
The main challenge of time-to-event prediction is that the observed outcomes are typically censored, meaning that the exact event time is unknown for some data due to early-end experiments or a lack of follow-up, making the estimation problem challenging. Conventional statistical machine learning techniques focus on the estimation of the hazard function, mathematically defined as the ratio of the probability density to the time-to-event function, representing the conditional probability density that a single nonrepeatable event will occur in a particular time interval, given that the item did not experience the event before that time. 

The Cox proportional hazards model (Cox model) \cite{cox1972regression}, proposed by Cox in 1972, offers a straightforward approach to handling censoring by assuming proportional hazards across covariates while leaving the baseline hazard function unspecified. Faraggi and Simon \cite{faraggi1995neural} introduced an extension of the Cox model by replacing its linear predictor with a one-hidden layer multilayer perceptron (MLP). Recent advancements of the Cox model with deep neural networks, e.g., DeepSurve \cite{katzman2018deepsurv} and Cox-CC \cite{kvamme2019time}, show promising performance. However, the proportional hazards assumption may not hold in complex scenarios, potentially leading to biased estimates and inaccurate predictions. To address this limitation, Kvamme et al. \cite{kvamme2019time} proposed a time-dependent Cox model to account for time-varying covariates. Furthermore, the Cox-based model estimates the baseline hazard function solely based on observed event times, which can introduce extra biases or information loss when data is limited. Random Survival Forests (RSF) \cite{ishwaran2008random}, a non-parametric model that builds upon the random forest algorithm and ensemble learning, shows advantages where traditional parametric or semi-parametric methods may not be suitable or when the underlying survival distribution is complex. In addition to estimating the time-to-event distribution, recent deep-learning advanced approaches also focus on improving prediction performance with new optimization strategies. For instance, DeepHit \cite{lee2018deephit}, a probability mass function-based discrete-time model, shows promising concordance index results with a loss function designed to improve event time ranking while disregarding the calibration of the predictions.

In this paper, we propose an evidence-based time-to-event prediction model that does not rely on specific forms of data distribution assumptions. Instead, we calculate the evidence of a time interval directly under the framework of belief functions \cite{dempster67,shafer1976mathematical} and random fuzzy sets \cite{denoeux2021belief,denoeux2023reasoning}. The proposed approach modifies the ENNreg model introduced in \cite{denoeux22,denoeux2023quantifying} to account for censored data. Prediction uncertainty is quantified using  Gaussian random fuzzy numbers (GRFNs) \cite{denoeux2023reasoning}, a newly introduced family of random fuzzy subsets of the real line.  In addition to providing the most plausible event time, our model outputs two additional quantities: standard deviation and precision measuring, respectively, aleatory and epistemic prediction uncertainties. The model is fitted by minimizing a generalized negative log-likelihood loss function.

The rest of this paper is organized as follows. Background notions are first recalled in Section \ref{sec:background}. The proposed model is then introduced in Section \ref{sec: methodology}, and experimental results are reported in Section \ref{sec: exp}. Finally, Section \ref{sec: conclu} concludes the paper.

\section{Background}
\label{sec:background}

%Data censoring can be regarded as imperfect data with only partial available knowledge, i.e., we only know the event does not happen at time $t$ and will occur after $t$, but we are ignorant about the exact event time. We, therefore, extend GRFNs into real-line time-to-event prediction tasks to address the censoring problem.

%\subsection{Time-to-event prediction}

%The \emph{survival function} $S(t|x)$ is used to identify the probability that an event will occur over time $t$ where $x$ is the vector of covariates; it is defined
%\begin{equation}
%    S(t|x)=P(T\ge t|x)=1-F(T<t|x),
%    \label{eq: S(t)}
%\end{equation}
%with $F(T<t|x)$ represents the probability that the event time is less than $t$.

%Denoting samples by $i$, with covariates $x_i$, and censoring indicator $D_i$, the likelihood of the event time $t_i$ is given by  
%\begin{equation}
%\begin{aligned}
%    L=\prod_{i}^{}f(t_i\mid x_i)^{D_i}S(t_i\mid x_i) ^{1-D_i},
%\end{aligned}
%\label{eq: nll_c}
%\end{equation}
%$L$ is referred to as \emph{full likelihood}. If the density function $f(t_i\mid x_i)$ is available, $S(t_i\mid x_i)$ can be calculated with Eq \eqref{eq: S(t)}, and the time-to-event prediction model can, therefore, be fit by maximizing the likelihood function \eqref{eq: nll_c}.

The theory of epistemic random fuzzy sets (RFSs) was introduced in \cite{denoeux2021belief,denoeux2023reasoning} as an extension of Dempster-Shafer theory allowing us to represent both partially reliable and vague evidence. In short, an RFS is a mapping from a probability space to the fuzzy powerset of another space, verifying some measurability property. The reader is referred to the cited references for a general exposition of this theory. Hereafter, we briefly recall the notions of Gaussian and lognormal random fuzzy numbers in Sections \ref{subsec:GRFN} and \ref{subsec:tGRFN}, respectively.

\subsection{Gaussian random fuzzy numbers}
\label{subsec:GRFN}
A Gaussian fuzzy number (GFN) is a fuzzy subset of the real line with membership function $x\mapsto \exp(-0.5h(x-m)^2)$, where $m\in \reels$ and $h\ge 0$ are the mode and precision parameters. A Gaussian random fuzzy number (GRFN) $\tT$ is an RFS defined as a GFN whose mode $M$  is a Gaussian random variable with mean $\mu$ and variance $\sigma^2$ \cite{denoeux2023reasoning}. It is then defined by three parameters $\mu$, $\sigma^2$ and $h$ and we write  $\tT \sim \tN(\mu, \sigma^2, h)$. The family of GRFNs is closed under the product-intersection rule, a combination operator generalizing Dempter's rule \cite{denoeux2023reasoning}. A GRFN defines a belief function of the real line. Formulas for the degrees of belief and plausibility of any real interval are given in \cite{denoeux2023reasoning}.

\subsection{Lognormal random fuzzy numbers}
\label{subsec:tGRFN}
A GRFN is a model of uncertainty about a variable taking values in the whole real line. In contrast, in time-to-event analysis, the response variable is positive. Uncertainty about such a variable is better represented by a lognormal random fuzzy number as introduced in \cite{denoeux2023parametric}. 

In general, let $\psi$ be a one-to-one mapping from $\reels$ to a subset $\Lambda\subseteq\reels$. Its extension $\tpsi$ maps each fuzzy subset $\tF$ of $\reels$ to a fuzzy subset $\tpsi(\tF)$ of $\Lambda$ with membership function $\lambda\mapsto \tF[\psi^{-1}(\lambda)]$. Let $[0,1]^\reels$ denote the set of all fuzzy subsets of $\reels$, and $\tY: \Omega \rightarrow [0,1]^\reels$ be a RFS. By composing $\tpsi$ with $\tY$, we obtain a new RFS $\tpsi\circ\tY: \Omega \rightarrow [0,1]^\Lambda$. For any event $C\subseteq \Lambda$, we have 
\begin{equation}
\label{eq:belpl}
Bel_{\tpsi\circ\tY}(C)=Bel_{\tY}(\psi^{-1}(C)) \quad \text{and} \quad  Pl_{\tpsi\circ\tX}(C)=Pl_{\tY}(\psi^{-1}(C)).
\end{equation}

Taking $\tY\sim \tN(\mu,\sigma^2,h)$, $\Lambda=[0,+\infty)$ and $\psi=\exp$, we obtain a \textit{lognormal random fuzzy number} $\tT:\widetilde{\exp}\circ \tY$ denoted by $\tT\sim T\tN(\mu,\sigma^2,h,\log)$. We can remark that $\widetilde{\log}\circ \tY\sim\tN(\mu,\sigma^2,h)$. Degrees of belief $Bel_{\tT}(I)$ and $Pl_{\tT}(I)$ for any interval $I\subseteq [0,+\infty)$ can easily be computed from \eqref{eq:belpl} and formulas given in \cite{denoeux2023reasoning} for GRFNs. %For any event $A\subseteq [0,+infty)$, \cite{denoeux2023parametric}.

\section{Model}
\label{sec: methodology}
%In this section

%The theory of epistemic random fuzzy sets \cite{} is a general theory of uncertainty encompassing both possibility theory \cite{} and the Dempster-Shafer theory of belief functions \cite{} as special cases.

%Within this framework, Gaussian random fuzzy numbers (GRFNs) have recently been introduced as a practical model of uncertainty about real variables \cite{}. However, the limited exibility of this model does not allow it to represent all kinds of beliefs encountered in applications. 

%Dempster-Shafer theory (DST) of evidence and possibility theory are two powerful mathematical frameworks for representing partial information and reasoning with uncertainty. DST makes it possible to represent partially reliable evidence and possibility theory allows us to express uncertainty associated with vague information such as concency by fuzzy sets. Encompassing both DST and possibility theory, GRFNs are introduced as a practical model to generate real line prediction as well as offering uncertainties about real variables. 

%For a give time $T$, if the observation is not censored 
%%%%%%%%%%%%%%%%%%%%%%%%%%%%%%%%%%%%%%%%%%%%%%%%%%%%%%%%%%%%%%%%%%%%%%%%%%%%%%%%%%%%%%%%%%%%%%%%%%%%%%%%%%%%%%%%%%%%%%%%%%%%%%

Our approach is based on the ENNreg model introduced in \cite{denoeux2023quantifying}. This model will be recalled in Section \ref{subsec:ENNreg}. The loss function adapted to censored data will then be described in Section \ref{subsec:loss}.

\subsection{Evidential time-to-event prediction network}
\label{subsec:ENNreg}

In time-to-event analysis, the response of an event time is always positive, while a GRFN is a model about a variable taking values in the whole real line. Following the idea of \emph{Lognormal random fuzzy numbers} as we introduced in Section \ref{subsec:tGRFN}, we construct a transformed GRFN-based evidential neural network to map predictions into the positive real timeline with $Y=\log(T)$, where $T$ is the time to event.
%We adopt the approach introduced in \cite{denoeux2023quantifying} and construct a GRFN-based evidential neural network to predict $Y=\log(T)$, where $T$ is the time to event. 
Here the network is composed of three layers: the distance layer, the evidence mapping layer, and the fusion layer. The distance  layer computes the  distances between the input vector $x$ and each prototype $p_k$ with a positive scale parameter  $\gamma_k$: $s_k(x)=\exp(-\gamma_k^2 \| x-p_k \|^2)$. 
For each prototype $p_k$, the evidence mapping computes a GRFN $\tN(\mu_k (x), \sigma_k^2, s_k(x)h_k)$, where $\sigma_k^2$ and $h_k$ are variance and precision parameters, and $\mu_k(x)$ is given by $\mu_k(x)=\beta_{k}^{T}x +\beta_{k0}$, where  $\beta_{k}$ is a $p$-dimensional vector of coefficients and $\beta_{k0}$ is a scalar parameter. %The quantity $\mu_k(x)$ and $\sigma^2_k$ can be seen as an estimation of the conditional expectation and variance, respectively, of the response $log(\tilde {T}_k(x))$ given that $x$ is close to $p_k$; $h_k$ is the precision.
The evidence fusion layer combines evidence from the $K$ prototypes using the unnormalized product-intersection combination rule $\boxplus$ \cite{denoeux2023reasoning} and outputs a final GRFN $\tY(x) \sim \tN(\mu(x),\sigma^2(x),h(x))$ given by 
 \begin{equation*}
\mu(x)=\frac{\sum_{k=1}^{K} s_k(x)h_k\mu_k(x)}{\sum_{k=1}^{K} s_k(x)h_k}, \quad
\sigma^2(x)=\frac{\sum_{k=1}^{K}s^2_k(x)h^2_k\sigma^2_k}{(\sum_{k=1}^{K} s_k(x)h_k)^2},
 \end{equation*}
 and 
 $
     h(x)=\sum_{k=1}^{K}s_k(x)h_k.
 $
Output  $\mu(x)$ denotes the most plausible time-to-event prediction, $\sigma^2(x)$ denotes the variance around $\mu(x)$ (aleatory uncertainty), and $h(x)$ denotes the precision of the prediction (epistemic uncertainty). Uncertainty about $T$ is then described by the lognormal RFN $\tT \sim T\tN(\mu(x),\sigma^2(x),h(x),\log)$.
%%%%%%%%%%%%%%%%%%%%%%%%%%%%%%%%%%%%%%%%%%%%%%%%%%%%%%%%%%%%%%%%%%%%%%%%%%%%%%%%%%%%%%%%%%%%%%%%%%%%%%%%%%%%%%%%%%%%%%%%%%%%%%
\subsection{Loss function}
\label{subsec:loss}

To optimize the proposed framework, we use negative generalized log-likelihood loss defined in \cite{denoeux2023quantifying}, and adapt it to account for both uncensored and censored data. 
If the data is not censored, the continuous event time $\tY$ is always observed with finite precision. Therefore, instead of observing an exact value, we actually observe an interval $[y]_{\epsilon}=[y-\epsilon, y+\epsilon]$ centered at $y$. Our prediction evidence can, therefore, be characterized by either the degree of belief $Bel([y]_{\epsilon} )$ or the plausibility $Pl([y]_{\epsilon})$. Conversely, if the data is censored, the event time $\tY$ will be observed in interval $[y,\infty)$. Our prediction evidence can now be represented as the degree of belief $Bel([y,\infty))$ or plausibility functions $Pl([y,\infty))$ in the time interval $[y,\infty)$. We can optimize the time-to-event function based either on $ L_{Bel}$ or $L_{Pl}$. While none of these two functions adequately measures the quality of the imprecise predictions as mentioned in \cite{denoeux2023quantifying}. %A detailed calculation regarding $Bel([y]_{\epsilon} )$ and $Pl([y]_{\epsilon})$ can be found in \cite{denoeux2023reasoning} by replacing the interval $[x, y]$ with $[y-\epsilon, y +\epsilon]$. 
%Following the suggestion of \cite{denoeux2023quantifying}, we consider the following weighted sum of $ L_{Bel}$ or $L_{Pl}$ with
Let $\tY$ be the output GRFN, $y=\log(t)$ the observation, and $D$ a binary censoring variable such that $D=1$ if $Y=y$, and $D=0$ if it is only known that $Y\ge y$. We, therefore, consider the following weighted sum of $ L_{Bel}$ or $L_{Pl}$ 
\begin{equation*}
     \calL_{\lambda,\epsilon}(\tY,y,D)= \lambda \overline{\calL}(\tY,y,D)+(1-\lambda) \underline{\calL}(\tY,y,D),
 \end{equation*}
with 
\begin{equation*}
    \overline{\calL}(\tY,y,D)=- D \ln Bel_{\tY}([y-\epsilon,y+\epsilon]) - (1-D) \ln Bel_{\tY}([y,\infty)),
\end{equation*}
and 
\begin{equation*}
    \underline{\calL}(\tY,y,D)=- D \ln Pl_{\tY}([y-\epsilon,y+\epsilon])- (1-D)\ln Pl_{\tY}([y,\infty)),
\end{equation*}
where $\lambda$ is the hyperparameter that controls the cautiousness of the prediction (the smaller, the more cautious). We set $\lambda=0.1$ to enable the model to focus more on plausibility optimization. The hyperparameter $\epsilon$ was set to $10^{-6}$ to present an infinitesimal time interval.
%The choose of hyperparameter $\lambda$ is related to the censoring rate  
%We used $\lambda=0.1$ and $\epsilon=10^{-6}$. 

The network is trained by minimizing the regularized average loss
\[
C_{\lambda,\epsilon,\xi,\rho}^{(R)}(\Psi)= \frac1n\sum_{i=1}^n \calL_{\lambda,\epsilon}(\tY(x_i;\Psi),y_i,D_i) + \frac{\xi}K \sum_{k=1}^K h_k + \frac{\rho}K \sum_{k=1}^K \gamma_k^2,
\]
where $\Psi$ is the vector of all parameters (prototypes are included as well) in the network, $\tY(x_i;\Psi)$ is the network output for input $x_i$, and $\xi$, $\rho$ are two regularization parameters. The first regularizer term has the effect of reducing the number of prototypes used for the prediction (e.g., setting $h_k = 0$ to discarding prototype $k$), while the second regularizer term shrinks the solution towards a linear model (e.g., setting $\gamma_k = 0$ for all $k$ yields a linear model)
In \cite{denoeux2023quantifying}, $\xi$ and $\rho$ are tuned by cross-validation. In the experiments reported in Section \ref{sec: exp}, we kept them fixed at $\xi=\rho=0.1$ for simplicity.
%-------------------------------------------------------------------------

%-------------------------------------------------------------------------
\section{Experimental results}
\label{sec: exp}
We will now show some qualitative results of our method applied to a simulated dataset with various data censoring scenarios (Section \ref{subsec:ex}), and compare its performance to state-of-the-art time-to-event prediction methods on two real-world datasets (Section \ref{subsec:real}).

\subsection{Illustrative example on simulated dataset}
\label{subsec:ex}
We first consider artificial data with the following distribution: the input $X$ has a uniform distribution in the interval $[-2, 2]$, and the response is
\begin{equation}
T=\exp \left[1.5 X+2 \cos (3 X)^{3}+\frac{X+5}{3 \sqrt{5}} V\right],
\end{equation}
where $V \sim N(0, 1)$ is a standard normal random variable. To simulate data censoring scenarios, two elements were incorporated: the event censoring state indicator $D$ and a random censoring value $C$. The event indicator $D$ has a Bernoulli distribution, denoted as $D \sim B(p)$, where $1-p$ represents the censoring rate (set as $0.1$ and $0.7$). For events flagged with a censoring indicator $D=1$, a negative value $C$ is added to $T$ to emulate right censoring. Here, the value $C$ follows a uniform distribution, with $C \sim U(-1, 0)$ and $C \sim U(-2, 0)$ representing different degrees of censoring severity. Learning and validation sets of size $n = 4000$ and $n = 1000$ were generated. 

The model was initialized with $K = 40$ prototypes. The targets $y=\log(t)$, the network outputs $\mu(x)$, along with belief prediction intervals (BPIs) at levels $\alpha \in \{0.5, 0.9, 0.99\}$ are shown in Fig. \ref{fig: censoring}. BPIs, as defined in \cite{denoeux2023quantifying}, are intervals centered at $\mu(x)$ with the degree of belief $\alpha$ to contain the true value of the response variable. 
%When only 10\% data are censored, our model can predict a time-to-event function (red line) closely aligns with the ground truth function (blue broken line), and the predicted BPIs can effectively encompass the majority of data points, as Figure \ref{fig: 10_[-1,0]} shows. When 70\% data are censored, as shown in Figure \ref{fig: 70_[-1,0]}, the predicted time-to-event function becomes smoother compared with the one in Figure \ref{fig: 10_[-1,0]} and has an upward bias relative to the true sample distribution as expected, while the predicted BPIs can still effectively encompass the majority of data points with a more width BPIs. When the censoring interval increases, e.g., from $[0, 1]$ to $[0, 2]$,  our model still shows good performance with a more wider BPIs. 
%. A BPI at level $\alpha$ is, thus, an interval that is believed to a degree $\alpha$ to contain the true value of the response variable. 
When only 10\% of the data are censored, our model predicts a time-to-event function (red line) that closely aligns with the ground truth function (blue broken line), and the predicted BPIs effectively encompass the majority of data points, as shown in Fig. \ref{fig: 10_[-1,0]}. With 70\% of the data censored (Fig. \ref{fig: 70_[-1,0]}), the predicted time-to-event function becomes smoother with fewer details and exhibits an upward bias relative to the true sample distribution, as expected. Nevertheless, the BPIs still effectively encompass the majority of data points, though they are wider. When the censoring interval increases, for example, from $[0, 1]$ (Fig. \ref{fig: 10_[-1,0]})  to $[0, 2]$ (Fig.\ref{fig: 10_[-2,0]}), our model continues to perform well with even wider BPIs.

We can conclude that for different data censoring scenarios, the predicted time-to-event functions closely model the actual regression function, even when the data is highly censored. Furthermore, the BPIs effectively encompass the majority of data points. These observations illustrate the robustness of our approach to varying data censoring conditions.
\iffalse
\begin{figure}[htb]
\centering
\includegraphics[width=\textwidth]{figures/simulated_example_2.pdf}
\caption{Simulated data, actual regression function (blue broken lines), and predictions obtained from the trained model with $K = 40$ prototypes. Predicted values $\log(y)$ are depicted by red solid lines, while belief prediction intervals (BPIs) at levels $\alpha \in \{0.5, 0.9, 0.99\}$ are represented by shaded areas in blue, green, and orange. The first and second rows are data with censor interval $[-1,0]$ and $[-2,0]$, respectively. The first, second, and third columns are data with 10\%, 40\%, and 70\% censoring rates.
\label{fig: censoring}}
\end{figure}
\fi

\begin{figure}[htb]
\subfloat[\label{fig: 10_[-1,0]}]{\includegraphics[width=0.5\textwidth]{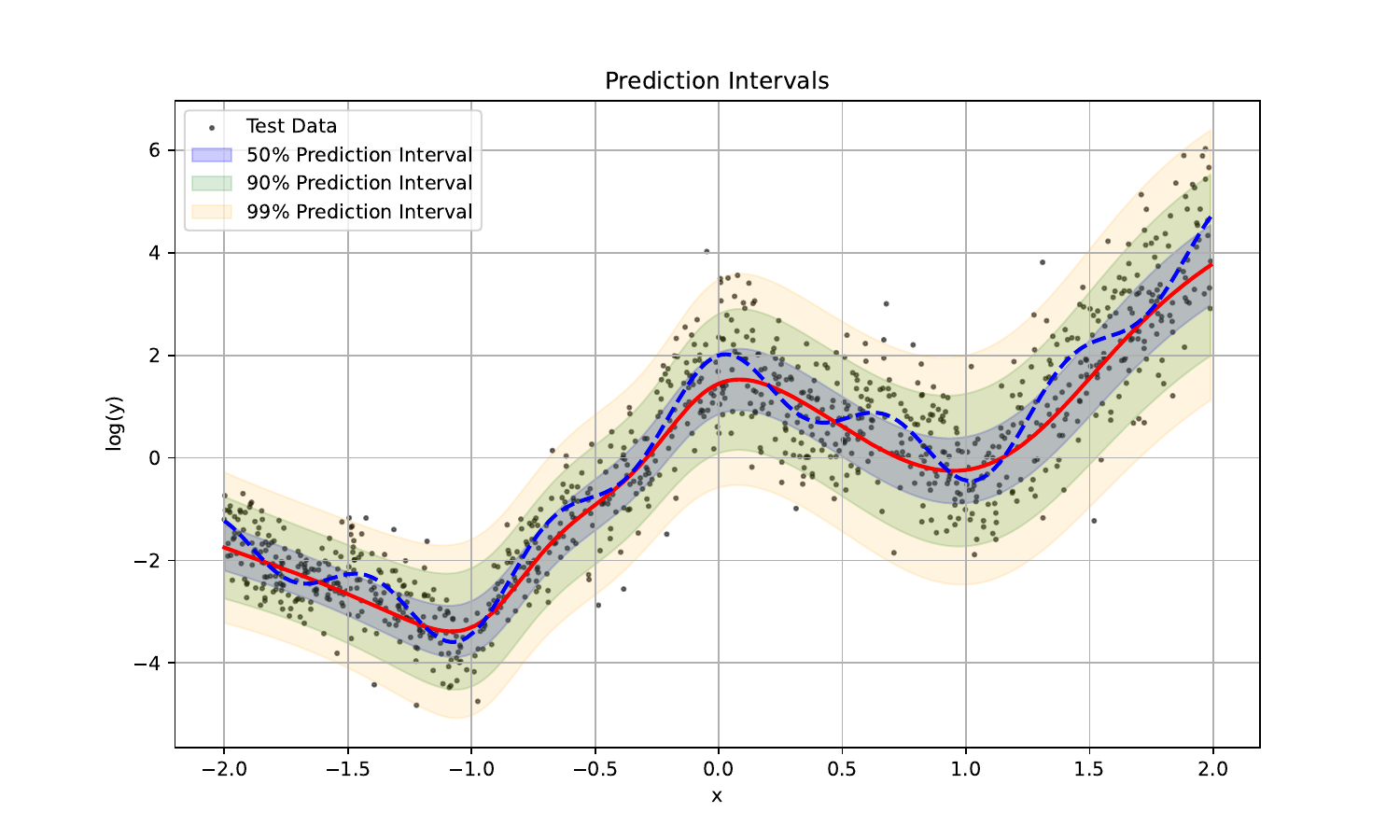}}
\subfloat[\label{fig: 70_[-1,0]}]{\includegraphics[width=0.5\textwidth]{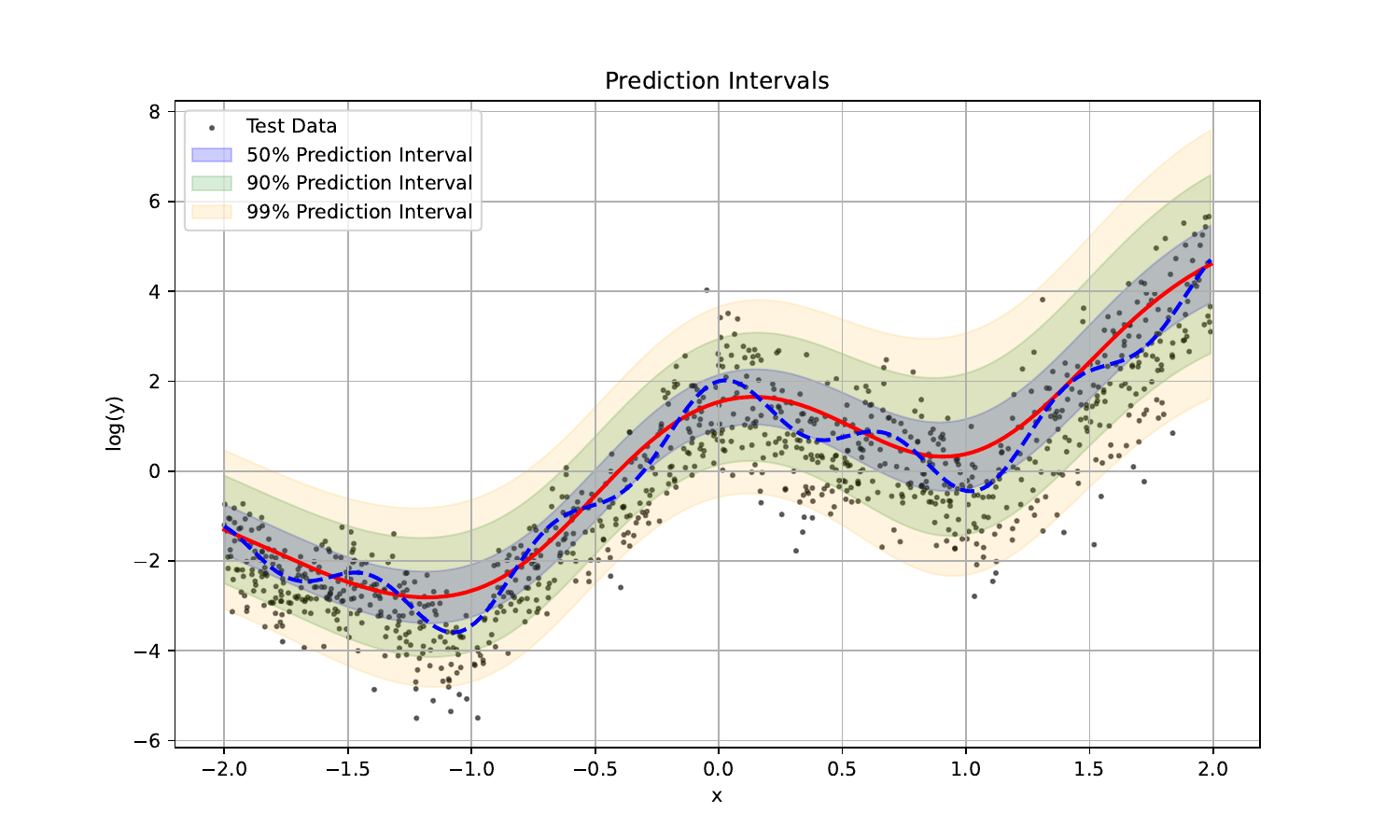}}\\
\subfloat[\label{fig: 10_[-2,0]}]{\includegraphics[width=0.5\textwidth]{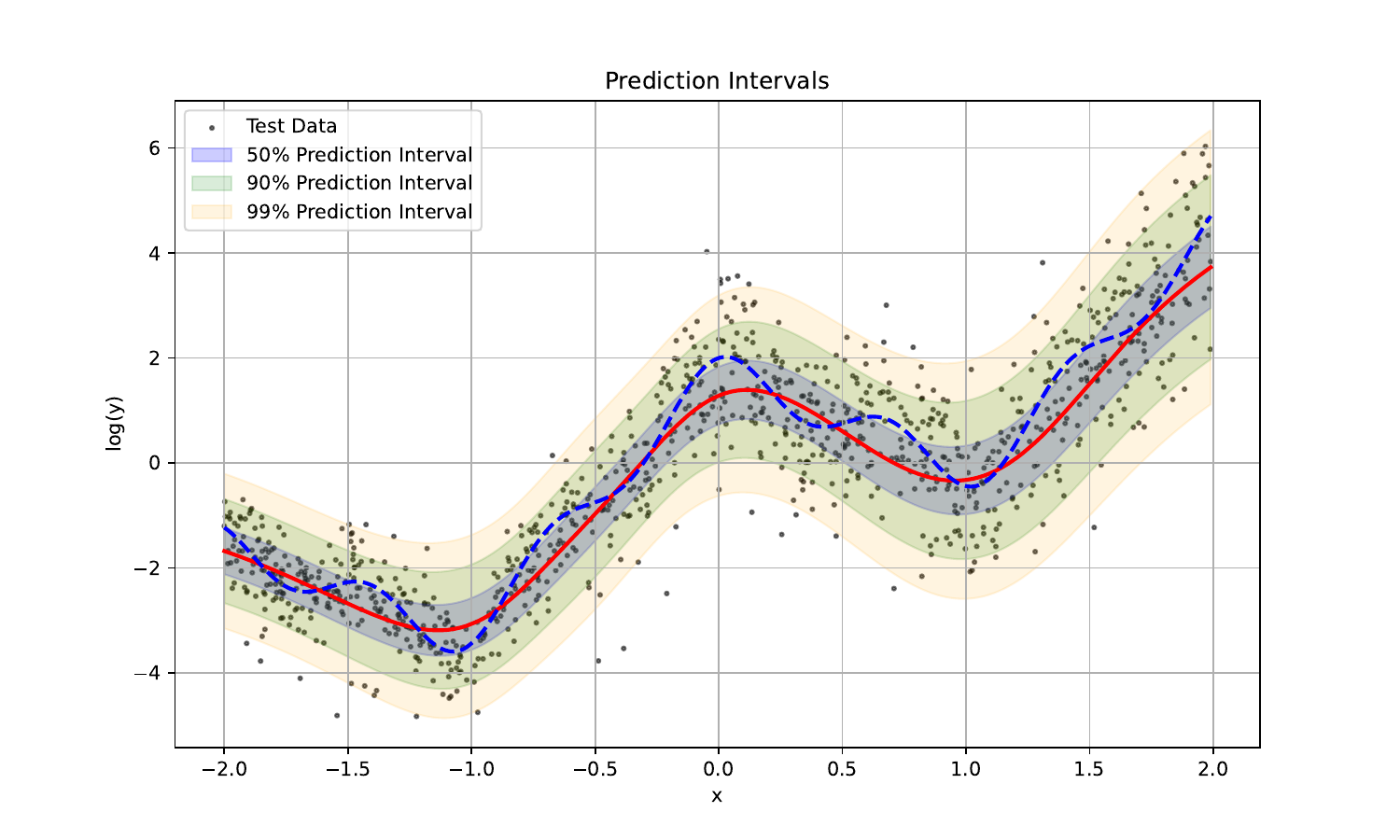}}
\subfloat[\label{fig: 70_[-2,0]}]{\includegraphics[width=0.5\textwidth]{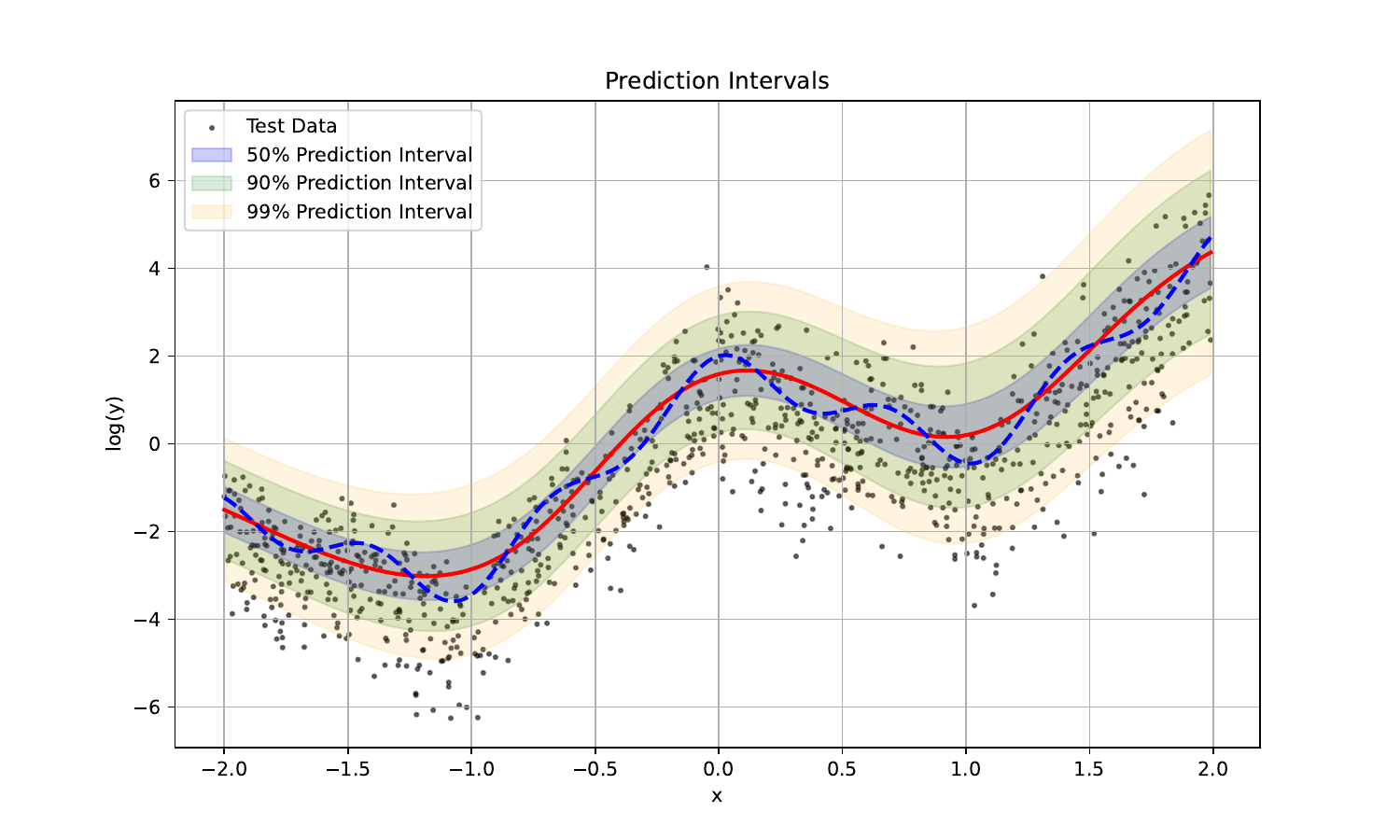}}
\caption{Simulated data, actual regression function (blue broken lines), and predictions obtained from the trained model. Predicted values $\log(y)$ are depicted by red solid lines, while belief prediction intervals (BPIs) at levels $\alpha \in \{0.5, 0.9, 0.99\}$ are represented by shaded areas in blue, green, and orange. The first and second rows are data with censoring intervals $[-1,0]$ and $[-2,0]$, respectively. The first and second columns are data with 10\% and 70\% censoring rates, respectively. \label{fig: censoring}}
\end{figure}

%\begin{figure}
%\begin{center}
%\subfloat[\label{fig:100}]{\includegraphics[width=0.5\textwidth]{figures/100.jpg}}
%\subfloat[\label{fig:90}]{\includegraphics[width=0.5\textwidth]{figures/90.jpg}}\\
%\subfloat[\label{fig:60}]{\includegraphics[width=0.5\textwidth]{figures/60.jpg}}
%\subfloat[\label{fig:30}]{\includegraphics[width=0.5\textwidth]{figures/30.jpg}}
%\caption{Prediction performance on different data censoring rates, (a) No data censoring, (b) 10\% data censoring, (c) 30\% data censoring and (d) 60\% data censoring. 
%\label{fig: censoring}}
%\end{center}
%\end{figure}

\subsection{Comparative results on real-world datasets}
\label{subsec:real}

We further evaluated our approach using two real-world time-to-event datasets provided by \cite{katzman2018deepsurv}. These are the Molecular Taxonomy of Breast Cancer International Consortium (METABRIC) dataset, comprising 1904 samples with a censoring rate of 42\%, and the Rotterdam Tumor Bank and German Breast Cancer Study Group (GBSG) dataset, containing 2232 samples with a censoring rate of 43\%. Following \cite{kvamme2019time}, we used the concordance index ($C_{idx}$) to evaluate the prediction performance, as well as the integrated Brier score (IBS) and the integrated binomial log-likelihood (IBLL) to evaluate the calibration of the estimates. We used five-fold cross-validation and repeated it five times. %It estimates the probability that, for a random pair of individuals, the predicted survival times of the two individuals have the same ordering as their true survival times. Furthermore, we also test the performance of \emph{Integrated Brier score (IBS)} and \emph{Integrated Binomial Log-Likelihood (BLL)} to measure both the discrimination and calibration of the estimations. A detailed description of the evaluation metrics can be found in \cite{kvamme2019time}.
We compared our methods to the baseline Cox method, RSF \cite{ishwaran2008random}, Deepsurv \cite{katzman2018deepsurv}, Cox-CC \cite{kvamme2019time}, Cox-Time \cite{kvamme2019time} and DeepHit \cite{lee2018deephit}. Hyperparameter values for the compared methods are given in the documentation of the Pycox package\footnote{\url{https://github.com/havakv/pycox}}.
%We compare our methods to a classical Cox regression model: Cox (Linear Cox model); a tree-based ensemble prediction model: Random Survival Forests (RSF); two versions of the Cox proportional hazards model incorporating neural networks: Deepsurv \cite{katzman2018deepsurv} and Cox-CC \cite{kvamme2019time}; a time-dependent relative risk model utilizing neural networks: Cox-Time \cite{kvamme2019time}; and a probability mass function-based discrete-time model with a loss for improved ranking: DeepHit \cite{lee2018deephit}. 

\begin{table}[htb]
\caption{Means and standard errors of $C_{idx}$, IBS and IBLL scores on the Metabric database for our method (ENNreg) and alternative methods. The best and second best results are, resp., in bold and underlined.}
\centering
\begin{tabular}{cc|c|c }
\hline
Methods &\multicolumn{1}{c}{$C_{idx}\uparrow$} & \multicolumn{1}{c}{IBS$\downarrow$}& \multicolumn{1}{c}{IBLL $\downarrow$ }\\
%\cline{2-7}
% &Metabric &GBSG&Metabric &GBSG&Metabric &GBSG\\  
\hline
Cox &0.633$\pm$9.3$\times10^{-3}$   &0.164$\pm$3.3$\times10^{-3}$ &0.497$\pm$1.1$\times10^{-2}$ \\
RFS &0.644$\pm$1.2$\times10^{-3}$   & 0.173$\pm$0.9$\times10^{-3}$   & 0.510$\pm$2.0$\times10^{-3}$  \\
Deepsurv &0.646$\pm$7.4$\times10^{-3}$  &\textbf{0.162}$\pm$3.6$\times10^{-3}$    &\underline{0.493}$\pm$1.2$\times10^{-2}$   \\
Cox-cc  &0.641$\pm$2.1$\times10^{-3}$ &\underline{0.163}$\pm$3.3$\times10^{-3}$   &\textbf{0.490}$\pm$8.6$\times10^{-3}$   \\
Cox-time&\underline{0.663}$\pm$1.0$\times10^{-2}$   &0.164$\pm$4.6$\times10^{-3}$    &0.488$\pm$1.1$\times10^{-2}$  \\
DeepHit &\textbf{0.672}$\pm$1.0$\times10^{-2}$  &0.173$\pm$2.6$\times10^{-3}$ &0.516$\pm$6.5$\times10^{-3}$    \\
ENNreg & \textbf{0.672}$\pm$9.4$\times10^{-3}$ & \underline{0.163}$\pm$2.1$\times10^{-3}$  & \textbf{0.490}$\pm$5.0$\times10^{-3}$ \\
          %BCESurv
\hline
\end{tabular}
\label{tab: result_Metabric}
\end{table}

\begin{table}[htb]
\caption{Means and standard errors of $C_{idx}$, IBS and IBLL scores on the GBSG database for our method (ENNreg) and alternative methods. The best and second best results are, resp.,  in bold and underlined.}
\centering
\begin{tabular}{cc|c|c }
\hline
Methods &\multicolumn{1}{c}{$C_{idx}\uparrow$} & \multicolumn{1}{c}{IBS$\downarrow$}& \multicolumn{1}{c}{IBLL $\downarrow$ }\\
%\cline{2-7}
%&Metabric &GBSG&Metabric &GBSG&Metabric &GBSG\\  
\hline
Cox  & 0.669$\pm$2.5$\times10^{-3}$  &\textbf{0.174}$\pm$3.3$\times10^{-3}$  &\underline{0.519}$\pm$1.7$\times10^{-3}$ \\
RFS   & 0.655$\pm$0.3$\times10^{-3}$  &0.190$\pm$0.5$\times10^{-3}$  &0.539$\pm$1.0$\times10^{-3}$ \\
Deepsurv & 0.666$\pm$8.4$\times10^{-3}$    & 0.180$\pm$1.9$\times10^{-3}$   & 0.531$\pm$5.1$\times10^{-3}$  \\
Cox-cc  & 0.672$\pm$3.3$\times10^{-3}$ & \textbf{0.174}$\pm$0.5$\times10^{-3}$  & 0.529$\pm$3.4$\times10^{-3}$  \\
Cox-time &\underline{0.678}$\pm$4.7$\times10^{-3}$   & \underline{0.177}$\pm$1.5$\times10^{-3}$   & 0.523$\pm$3.7$\times10^{-3}$  \\
DeepHit  &\underline{0.678}$\pm$4.5$\times10^{-3}$  & 0.195$\pm$1.0$\times10^{-3}$  & 0.565$\pm$2.6$\times10^{-3}$  \\
ENNreg  &\textbf{0.681}$\pm$2.2$\times10^{-3}$ &\textbf{0.174}$\pm$1.1$\times10^{-3}$  &\textbf{0.518}$\pm$2.8$\times10^{-3}$\\
          %BCESurv
\hline
\end{tabular}
\label{tab: result_GBSG}
\end{table}

Results for the two datasets are reported in Tables \ref{tab: result_Metabric} and \ref{tab: result_GBSG}. We can remark that all the compared methods are based on specific assumptions, and it is not surprising that some of them perform quite well for specific data distribution. The Cox model performs rather poorly, which was expected as it is based on very restrictive model assumptions. The methods that assume proportional hazards without linearity assumption, i.e., Cox-CC and DeepSurv perform worse, in general, than the less restrictive methods, namely, RFS and Deephit. 

Our ENNreg method achieved the best performance according to the $C_{idx}$ and IBLL criteria, and the second best. Notably, our proposal outperforms the continuous Cox-time model by a large amount and performs slightly better than the discrete DeepHit model. This result is interesting considering that we did not develop a time-dependent prediction model like Cox-time, nor did we use concordance for hyperparameter tuning as in Deephit. As we can see from the IBS and IBLL results, the promising concordance performance of Deephit comes at the cost of poorly calibrated survival estimates. In contrast, our proposal exhibits good calibration properties, with statistically significant differences observed in calibrated survival estimates for both datasets. We can, therefore, conclude that our evidence-based time-to-event prediction model,  based on minimal assumptions, demonstrates greater flexibility and robustness compared to state-of-the-art models that rely on restrictive hypotheses such as the proportional hazard assumption. %Aside from accommodating various data distributions, our method can effectively manage different data censoring scenarios, which is confirmed by Fig. \ref{fig: censoring} as well. 

\section{Conclusion}
\label{sec: conclu}
In time-to-event analysis, some proportion of the data is usually censored. In this paper, we have adapted the ENNreg model introduced in \cite{denoeux2023quantifying} to account for censored data, and applied it to time-to-event prediction. The model is trained using the logarithm of the response variable $T$ as the target variable and outputs a GRFN. Prediction uncertainty is, thus, quantified by a lognormal random fuzzy number, from which degrees of belief and plausibility of various events can be straightforwardly computed. In this paper, we have focused on prediction accuracy and calibration (assessed using standard performance criteria) and showed the good performance of our model on two datasets as compared to the state-of-the-art. In the future, we will further explore the advantages of uncertainty quantification in time-to-event tasks using belief functions, e.g., studying the standard deviation and precision of the prediction. We will also extend the comparison with state-of-the-art to a wider range of clinical medical datasets for different time-to-event tasks. Moreover, the study of mixtures of GFRNs to fit applications, e.g., finance analysis, should also be interesting.

\section{Acknowledgment}
This research is supported by A*STAR, CISCO Systems (USA) Pte. Ltd, and National University of Singapore under its Cisco-NUS Accelerated Digital Economy Corporate Laboratory (Award I21001E0002) and the National Research Foundation Singapore under AI Singapore Programme (Award AISG-GC-2019-001-2B).
%
% ---s o addres- Bibliography ----
 \bibliographystyle{splncs04}
 \bibliography{mybibliography}

\iffalse

\fi
\end{document}